\documentclass[letterpaper, 10 pt, conference]{ieeeconf}  %

\IEEEoverridecommandlockouts                              %

\usepackage{latexsym}
\usepackage{amsmath}
\usepackage{amssymb}
\usepackage{epsfig}
\usepackage{bbm}
\usepackage{caption}
\usepackage{graphicx,float}
\usepackage{cuted}
\usepackage{pifont}
\usepackage{makecell}
\usepackage{todonotes}
\usepackage{multirow}
\usepackage{color, colortbl}

\usepackage{cite}

\usepackage[ruled]{algorithm2e}

\usepackage{verbatim}
\usepackage{hyperref}
\hypersetup{%
  breaklinks,
  bookmarksnumbered=true,
  bookmarksopen=true,
  colorlinks=true,
  linkcolor=black,
  urlcolor=black,
  citecolor=black
}
\usepackage{xcolor}
\usepackage{subfiles}

\newcommand{\SystemName}{\texttt{SALON}}

\begin{document}
\bstctlcite{IEEEexample:BSTcontrol}

\title{SALON: Self-supervised Adaptive Learning for Off-road Navigation}

\author{Matthew Sivaprakasam$^{1}$, Samuel Triest$^{1}$, Cherie Ho$^{1}$, Shubhra Aich$^{1}$,\\ Jeric Lew$^{2}$, Isaiah Adu$^{3}$, Wenshan Wang$^{1}$, and Sebastian Scherer$^{1}$ %
\thanks{* This work was supported by ARL awards \#W911NF1820218 and \#W911NF20S0005.}%
\thanks{$^{1}$ Robotics Institute, Carnegie Mellon University, msivapra,striest,cherieh,saich,wenshanw,basti@andrew.cmu.edu}%
\thanks{$^{2}$ National University of Singapore, jericlew@u.nus.edu}
\thanks{$^{3}$ Pennsylvania State University, ioa5099@psu.edu}%
}

\maketitle

\IEEEpeerreviewmaketitle

\begin{abstract}
Autonomous robot navigation in off-road environments presents a number of challenges due to its lack of structure, making it difficult to handcraft robust heuristics for diverse scenarios. While learned methods using hand labels or self-supervised data improve generalizability, they often require a tremendous amount of data and can be vulnerable to domain shifts. To improve generalization in novel environments, recent works have incorporated adaptation and self-supervision to develop autonomous systems that can learn from their own experiences online. However, current works often rely on significant prior data, for example minutes of human teleoperation data for each terrain type, which is difficult to scale with more environments and robots. To address these limitations, we propose $\SystemName$, a perception-action framework for \textit{fast} adaptation of traversability estimates with \textit{minimal} human input. $\SystemName$ rapidly learns online from experience while avoiding out of distribution terrains to produce adaptive and risk-aware cost and speed maps. Within \textit{seconds} of collected experience, our results demonstrate comparable navigation performance over kilometer-scale courses in diverse off-road terrain as methods trained on 100-1000x more data. We additionally show promising results on significantly different robots in different environments. Our code is available at \href{https://theairlab.org/SALON/}{https://theairlab.org/SALON}

\end{abstract}

\section{Introduction}

Off-road autonomous driving is becoming an increasingly researched topic due to its wide range of applications. Robots are already being deployed in fields such as agriculture \cite{yepezponce2023robotfarming}, infrastructure monitoring \cite{gibb2018nondestructive}, and defense \cite{naranjo2016autonomous}, where they must operate in unstructured and diverse environments. To perform reliably, they must reason about terrain lacking clear structures and markings to navigate from one goal to another without crashing or getting stuck.

\begin{figure}[]
	\centering
	\includegraphics[width=.99\linewidth]{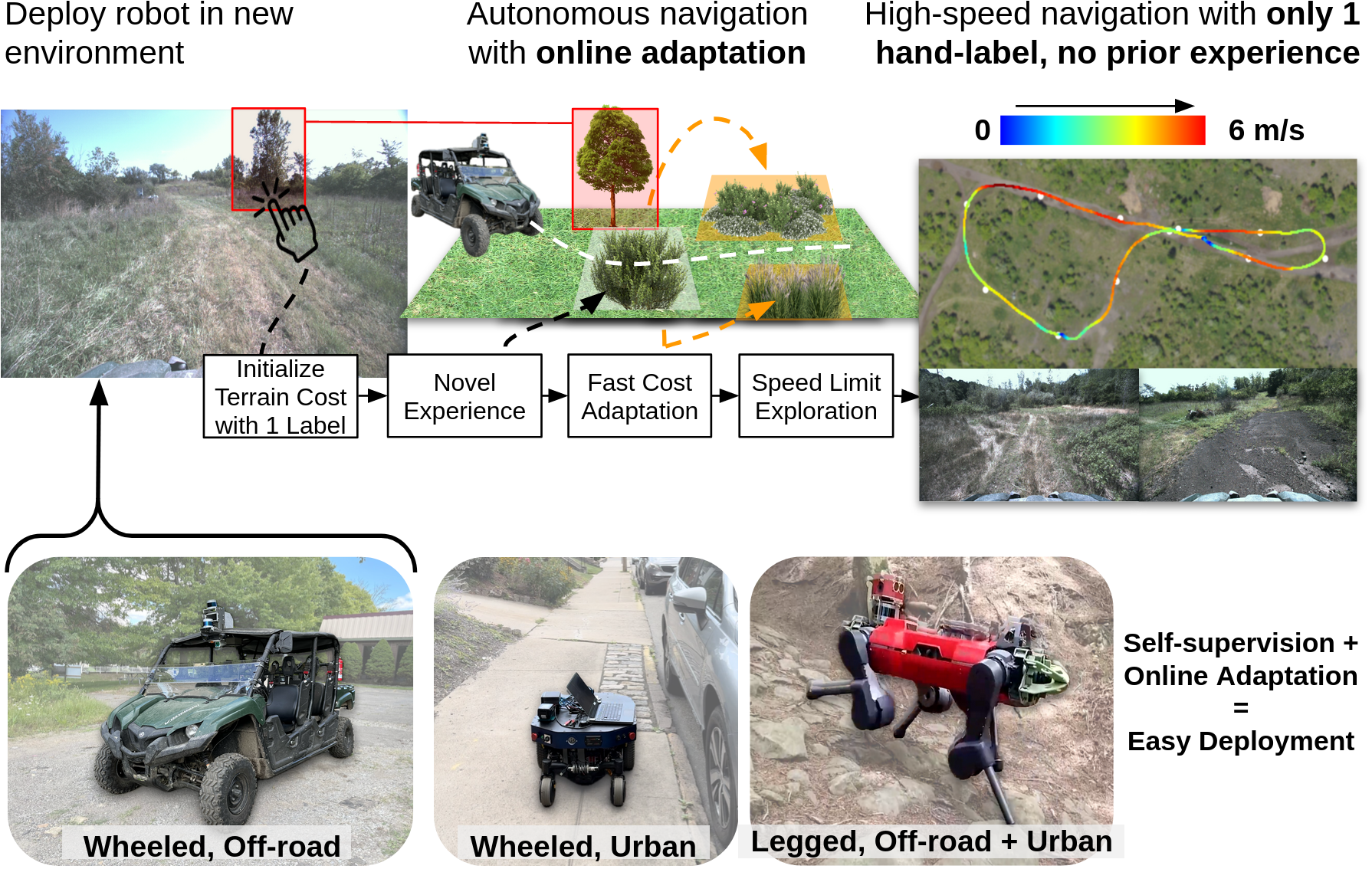}
	\caption{We present SALON, a framework for off-road navigation with no prior experience. With one prior hand-label, our system running \SystemName\ learns from its own experience in the real-world to predict where and how fast to drive.}
	\label{fig:title}
\end{figure}

Recent research has focused on improving navigation by generating costmaps with fine details, which captures the complexities of off-road terrain. For instance, detecting obstacles like rocks and trees without confusing them with traversable terrain such as small bushes and tall grass illustrates the level of detail required for effective navigation. Costmap generation methods, such as height-thresholding \cite{occupancy} and semantic segmentation \cite{maturana, pmlr-v164-shaban22a} often struggle in off-road domain due to basic assumptions. More expressive geometric analyses \cite{fankhauser, step, krusi, NOREN2025101034} reduce this limitation, but can require extensive hand-tuning and still fail to distinguish different terrain with similar geometry. Learned methods have shown potential to address these issues \cite{triest2023learning, 10160856, meng2023terrainnet, frey2024roadrunnerlearningtraversability, triest2024velociraptor} but are trained offline and struggle when deployed in new environments. 

To improve generalization in novel environments, several works incorporate adaptation in a self-supervised manner, developing proprioceptive signals that enable the system to learn from its own experiences online \cite{mattamala2024wildvisualnavigationfast, chen2023learningonthedrive, sathyamoorthy2022terrapn}. 
\textit{ALTER} \cite{chen2023learningonthedrive} adapts its visual traversability model using explicitly generated cost (e.g. ideal to lethal) from LiDAR-built geometric maps, but is limited to accurate range sensors, and does not link to autonomy. Most relevant to our work, WVN \cite{mattamala2024wildvisualnavigationfast} is a concurrent perception-action framework that learns traversability online by observing expert demonstration, then autonomously exploring within familiar terrain to refine its predictions. However, their anomaly detection severely penalizes areas where it has not traversed before, requiring teleoperation in all terrain types the robot should visit. This reliance on human teleoperation is not scalable when deploying in unknown environments, as the process may need to be repeated for different environments and robots. Such methods are insufficient for achieving \textit{fast adaptation in new environments with minimal human input}.

 We therefore propose $\SystemName$, a perception-action framework for fast adaptation of traversability estimates with minimal human input (only one click on an image, instead of minutes of expert teleoperation in all terrains). $\SystemName$ achieves this with the following design choices:
 \begin{itemize}
     \item \textbf{Mapping and learning in map space}: We use a mapping pipeline to project visual features from the camera into a Birds-Eye-View (BEV) map and associate the traversed cells with a cost and speed experienced by the robot. This results in cleaner, more distinctive maps.
     \item \textbf{One-shot cost augmentation}: a user simply clicks areas to avoid on prior images to initalize the system
     \item \textbf{Explicit Out-of-distribution (OOD) detection}: allows cleaner detection of anomalous objects, without long prior teleoperation in all safe terrain
     \item \textbf{Adaptive and risk-aware cost and speed maps}: Allows the robot to optimize mission-relevant metrics with novel experience
 \end{itemize}

Overall we present three contributions:
\begin{enumerate}
\item $\SystemName$, a novel adaptive perception-action framework to generate \textbf{adaptive costmaps and speedmaps}, allowing autonomous navigation with minimal interventions, given as few as one hand-label.
\item \textbf{Real-world experiments} demonstrating performance at a similar level as state-of-the-art methods trained offline on 100-1000x more data.
\item Qualitative results on \textbf{multiple heterogeneous robots} differing in terms of dynamics, sensors, cost functions, visual back-ends, and environments.
\end{enumerate}
Our code is open source, leveraging an existing mapping framework in order to facilitate deployment on other robots.

\section{Related Work}
\subsection{Costmap Generation for Off-road Driving}
There exists a large number of existing works on learning self-supervised costmaps, both on and off-road. Some approaches leverage privileged information to supervise neural networks that predict map information at a given timestep \cite{meng2023terrainnet, frey2024roadrunnerlearningtraversability, triest2024unrealnetlearninguncertaintyawarenavigation, chen2023learningonthedrive}, but this information consists of semantic segmentation or comes from handcrafted cost functions, both of which require copious amount of hand labels and tuning. Some methods aim to circumvent this explicit hand-labeling requirement by using expert demonstration data \cite{triest2023learning, wulfmeier}, and while the supervision comes from the data collection process itself new challenges arise with ensuring demonstration quality and adapting to novel stimuli without a human in the loop. Recent works, taking inspiration from older methods \cite{Angelova2007LearningAP, dupont, fankhauser}, have explored the potential for proprioception as supervision as it allows for a strong robot-specific relationship between experience and cost. Some of these methods leverage signals such as residuals between planned and expected trajectories \cite{cai2024evoradeepevidentialtraversability, matthew} or IMU-based roughness score \cite{10160856, pokhrel2024cahsorcompetenceawarehighspeedoffroad, rca}. However, above methods require a significant amount of training data. 
While recent works have leveraged pre-trained models or visual foundation models (VFMs) to reduce the amount of labeled inputs \cite{jung2024vstrongvisualselfsupervisedtraversability, triest2024velociraptor}, costmap generation methods that rely on statically-trained models may fail when deployed in out-of-distribution environments.

\subsection{Adaptive Methods for Costmap Generation}
To improve performance when deployed in new environments, many works incorporate adaptation in their autonomy stack to learn from online experience. However, previous works do not sufficiently achieve \textit{fast adaptation in new environments with minimal human input} due to insufficiently expressive features, sensing restrictions, and need for minutes of pretraining data before effective adaptation. We find works that adapt online with LiDAR geometry measurements \cite{chen2023learningonthedrive, seo2024metaversemetalearningtraversabilitycost, dahlkamp2006self, Bagnell2010learningnavigation, Sofman2006ImprovingRN}, but still struggle to reason about complex details present in natural environments. Instead, we propose a vision-based framework that is flexible to learn from different cost functions based on various sensors. We also find older works that learn from stereo camera measurements \cite{hadsell2008deep,Happold2006terrain}, but their use of less expressive image features hinders the prediction performance. 
More recently, adaptive methods have adopted the use of more expressive deep-learning based image features \cite{sathyamoorthy2022terrapn, mattamala2024wildvisualnavigationfast} to learn odometry-based cost, similar to our work. However, TerraPN \cite{sathyamoorthy2022terrapn} takes $\sim$25 minutes to learn which does not fulfill our need for fast adaptation, whereas ours can learn in the order of seconds of traversing a new type of terrain. Most relevant to our work, WVN \cite{mattamala2024wildvisualnavigationfast} had success taking it a step further using Vision Transformer-based foundation model features to achieve faster, more generalizable adaptation. However, their framework requires human teleoperation in the different types of terrains the robot is expected to visit. Such need for human teleoperation is not scalable, as human teleoperation may be needed for each combination of environments and robots. In contrast, our method can deploy and explore new types of terrain with minimal human input (only one click on an image, instead of minutes of human teleoperation in all expected terrains).

\begin{figure*}[]
	\centering
	\includegraphics[width=.8\linewidth]{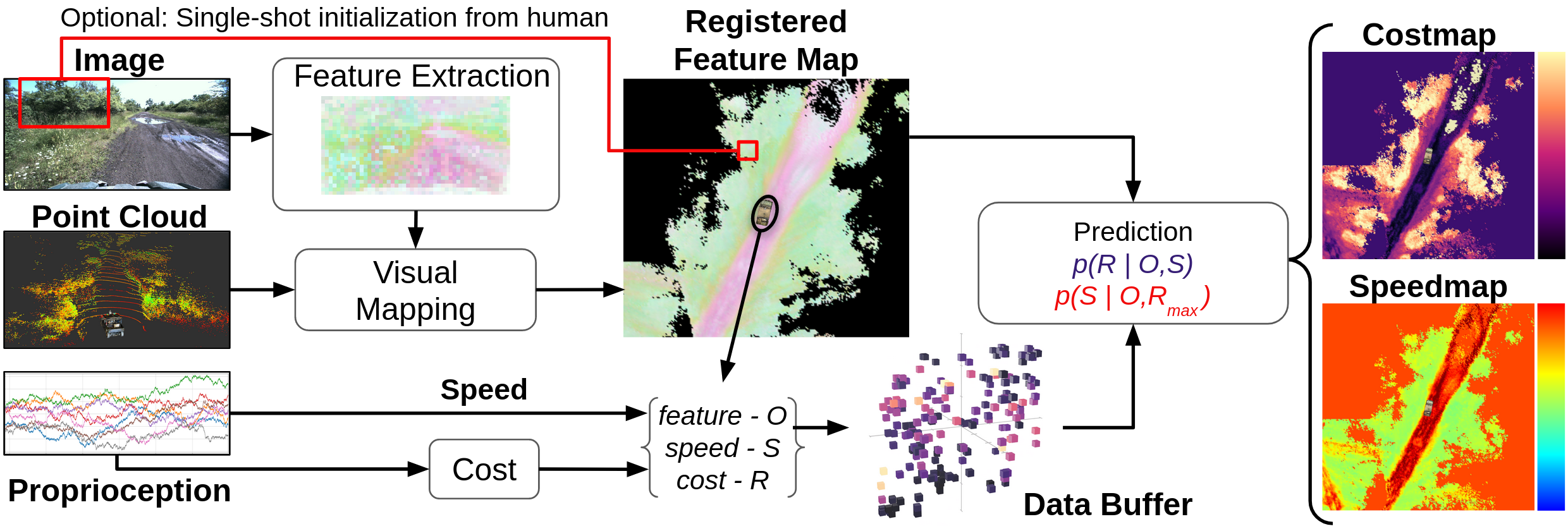}
	\caption{
    $\SystemName$ Overview: We learn rapidly from online experience with minimal human input to predict cost maps and speed maps.Visual foundation model (VFM) features in the map space, proprioceptive supervision, and smart data strategies together enable perception that adapts quickly to its environment.
 }
	\label{fig:flowchart}
\end{figure*}

\section{Adapting Perception with Online Experience}
Our approach is powered by generalizable features, self-supervised cost signals, intelligent data management, and probabalistic traversability estimation (Fig. \ref{fig:flowchart}). Together, they enable our system to adapt quickly to novel terrain without extensive prior demonstrations.

\subsection{Visual Mapping}
We leverage expressive features from VFMs aggregated into a BEV map as our terrain representation which is in turn used for prediction.

\subsubsection{Visual Backbone}
Pixel-level features in the image frame can be computed via a visual feature extractor, providing the mapping:

\begin{equation}
f_\theta(I_{3\times H \times W}) = D \in \mathbb{R}^{C \times H \times W}
\end{equation}

where $I$ is an RGB image, $D$ is a``featurized" image, with $C$ feature channels generated by feature extractor $f_\theta$. 

We use VFMs as feature extractors as we observe qualitatively that they not only can reason about commonly encountered terrain but also previously unexperienced terrain and objects. While the spatial resolution of these features is often lower due to model architecture, we find their powerful generalizability compensates for this shortcoming.

\subsubsection{Dimensionality Reduction}

VFMs produce features with hundreds of channels per pixel, which can be intractable to represent in BEV in real-time. To compress them into a lower-dimensional space, we first generate feature images $D$ from a subset of training images, then randomly sample pixel-level embeddings $d \in \mathbb{R}^C$ from each image. Inspired by vector of locally aggregated descriptors (VLAD), which is popular in the place-recognition domain \cite{vlad}, we perform K-Means clustering on the training subset to generate $k$ feature clusters ($F_{1:k}$). A k-dimensional descriptor is then generated for each VFM pixel $d$, where the $k$-th feature in the descriptor is the $L_1$ distance of $d$ to cluster center $F_k$ (Equation \ref{eq:vlad_features}).

\begin{equation}
\label{eq:vlad_features}
d_{VLAD}[k] = \| d - F_k \|_1
\end{equation}

In line with prior work \cite{keetha2023anyloc}, we find that this dimensionality reduction technique preserves semantic properties, which can provide strong priors for downstream tasks.

\subsubsection{BEV Mapping}
We aggregate these reduced FPV features in a BEV map using the same mapping method used in Velociraptor \cite{triest2024velociraptor}. Using known calibration, visual features from a camera are associated with 3D points from a lidar, then projected into a BEV map which is aggregated over time by applying an exponential moving average.

\subsection{Curation of the Self-Supervised Signal}

For a robot to adapt from its own experience, it requires a signal that associates different types of terrains with different costs in a way that matches human intuition. Following work by Castro et al. \cite{10160856}, we take inspiration from prior approaches \cite{dupont, stavens2012selfsupervisedterrainroughnessestimator, wellhausen} and use bandpower $BP$ of a signal across a frequency range $[f_{min}, f_{max}]$ as a way to compute roughness. While many works \cite{10160856, seo2024metaversemetalearningtraversabilitycost} compute bandpower for $Z$-axis (vertical) acceleration data alone, we find that for our full-scale system it is beneficial to include the other axes as well as readings from our vehicle supsension (henceforth referred to as shock travel).

In order to design a mapping from these sensor measurements to cost, we first collect a small dataset driving over different types of terrain at different speeds, periodically annotated by a passenger in the vehicle with a traversability score in the range 0-1. The roughness is generated from computing the bandpower for all proprioceptive signals $R = \sum_{i\in [a_x,a_y,a_z,shock...]} w_iBP(s_i,f^{min}_i, f^{max}_i) 
$,
where $w$ is the weight for each signal, and $s$ is the window length of data. In order to obtain the best set of parameters $[w_i,s_i,f^{min}_i, f^{max}_i]$ we optimize them to produce a roughness that matches the human annotations based on cumulative L1 error.

\subsection{Intelligent Data Maintenance}
Leveraging the visual BEV mapping described above as a perceptual representation, we have the robot store experience as it drives, where a sample collected at time $t$ contains:

\begin{enumerate}
	\item $O_t$ - The observed visual feature from the BEV map under the vehicle tire at time $t$
	\item $S_t$ - The speed that the vehicle was traveling
	\item $R_t$ - The roughness that the vehicle experienced
\end{enumerate}

Over time, the robot must throw out old samples to make room for new experiences. Rather than adopting a "first in, first out" (FIFO) strategy that can lead to catastrophic forgetting, we implement a strategy that aims to ensure an even distribution of data across the feature space. The VLAD features in the BEV representation by nature describe distance of observations to pre-defined clusters. Rather than throwing out the oldest sample, we leverage the semantic nature of our feature representation to instead throw out a sample $n_{\bar{C}\bar{S}}$ corresponding to the most common "semantic class" $\bar{C}$ and speed $\bar{S}$.

To verify this strategy, we compute the average pairwise distance between all points in the buffer for multiple sample trajectories, shown in Table \ref{tab:buffer_metrics} where the higher values for our method indicate a better coverage of the sample space in all of three different scenarios.

\begin{table}[b]
    \caption{Avg. Pairwise Distance Between Points in Buffer}
    \centering
    \begin{tabular}{|c||c|c|c|}
        \hline        
        Strategy & Scenario 1 & Scenario 2 & Scenario 3\\
        \hline
        FIFO &  3.47 &  3.60 & 3.59\\
        Remove $n_{\bar{C}\bar{S}}$ & \textbf{4.79} & \textbf{4.43} & \textbf{4.33}\\
        \hline
    \end{tabular}
    \label{tab:buffer_metrics}
\end{table}

\subsection{Costmap and Speedmap Estimation}
Given prior experience pairing roughness with visual features and speeds, the system needs a means of reasoning about the cost of the terrain ahead of it. We can predict the mean roughness $\mu_R$ and variance $v_R$ of a cell in the BEV map given its feature and the speed of the vehicle.

\begin{equation}
\mu_R, v_R = p(R | O, S)
\end{equation}

The mean and variance can be computed using Gaussian Process Regression (GPR) using a radial basis function (RBF) kernel. Note that this can also be approximated with a simple MLP network, but we find that in our case GPR allows faster adaptation with stable predictions.

Note that this formulation also provides an estimate of the variance, which in turn means that the final roughness prediction can be tuned using Conditional Value at Risk (CVaR) based on the user's preferred risk tolerance similar to other works \cite{step, cai2024evoradeepevidentialtraversability, triest2023learning}. The risk-adjusted predicted roughness assuming a Gaussian distribution becomes:

\begin{equation}
R = \mu_R + v_R \frac{\phi(\Phi^{-1}(\alpha_R))}{1-\alpha_R}
\end{equation}
Where $\alpha_R$ is set by the user to vary risk-tolerance.

The experience buffer can also be similarly leveraged to predict speedmaps that dictate the upper-bound speed that the robot should travel for each cell. The user can specify a maximum desired roughness $R_{max}$, and use the same method to instead predict the mean speed $\mu_S$ and variance $v_S$:

\begin{equation}
\mu_S, v_S = p(S|\ O, R=R_{max}),\  0 \leq R_{max} \leq 1
\end{equation}

which can in turn be used to compute a speed limit for the downstream controller. 

An issue arises here with out-of-distribution situations, where we are unlikely to obtain high-speed predictions without first experiencing high-speed data. We account for this by using CVaR with parameter $\alpha_S$ similarly to $\alpha_R$, but dynamically adapting it instead of having it set by the user. While the vehicle is traveling within some margin of the speed limit but experiencing roughness that is significantly less than the expected roughness $R_{max}$, $\alpha_S$ is incrementally increased, and decreased if it is exceeding $R_{max}$. This allows the robot to explore higher speeds until it obtains evidence that supports its predictions, and adjust its limits if encounters previously unseen terrain that is too rough.

\begin{figure}[]
	\centering
	\includegraphics[width=.99\linewidth]{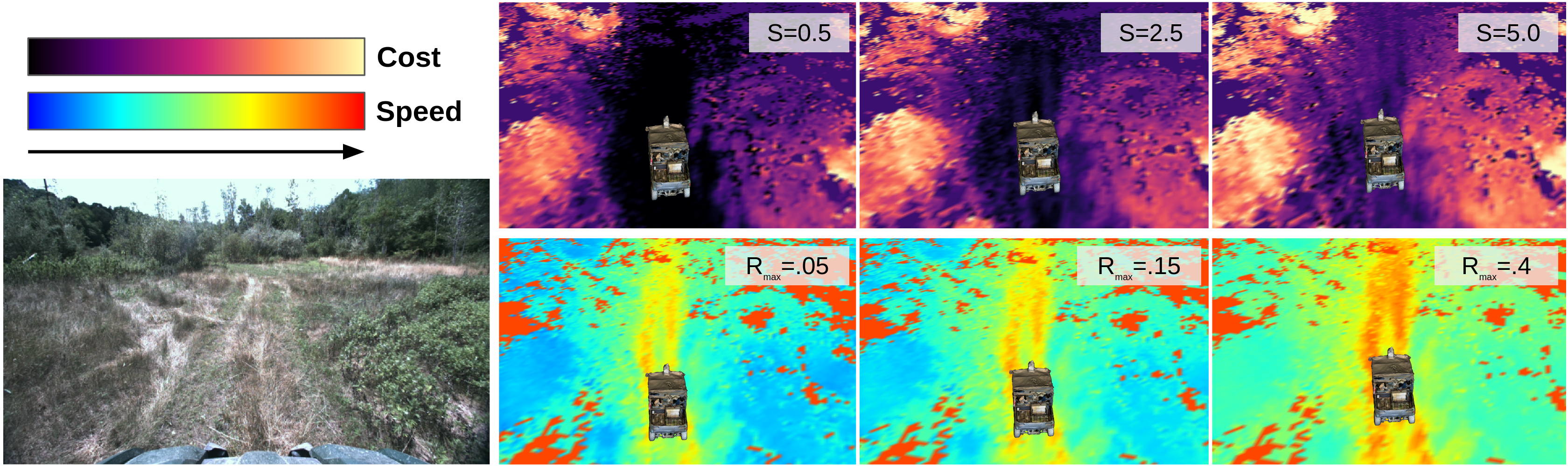}
     \caption{Our system understands the relationship between velocity and traversability, and will command speeds that match the user's tolerance for experienced cost. \textit{Top row: }costmaps, conditioned on different speeds increasing from left to right; \textit{Bottom row:} speedmaps, with a user-set maximum cost threshold increasing from left to right.}
	\label{fig:adaptation_title_horizontal}
\end{figure}

\subsection{One-Shot Costmap Augmentation}
Above, we show how a system can leverage features from VFMs to quickly learn costmaps and speedmaps without any human labels and adapt them with online experience. However, this formulation using signals such as roughness comes with the limitation that the robot can only learn about what it can physically drive on. This means reasonable predictions cannot be guaranteed for features that correspond to lethal objects.

We address this problem by relaxing the assertion of zero human labels to one human label for commonly-experienced lethal objects, such as trees. We find that by simply choosing a single feature from the BEV map that corresponds to a tree and permanently associating it with high roughness in the buffer, we can obtain high cost values for all the trees that the robot experiences, without needing to train the network further or have a user label several trees.

\subsubsection{Avoiding Out-of-Distribution Terrain}
While the one-shot costmap augmentation is effective for frequently-encountered lethal terrain like trees, it may not be possible for a user to label all unique types of lethal terrain. In order to avoid terrain such as foreign objects, we leverage the same uncertainty estimation presented in Velociraptor, where a map cell is considered lethal if its distance to all feature clusters exceeds a threshold. We apply an additional morphological erosion and dilation to the uncertainty estimate in the BEV space to remove noise.

\section{Experiments and Results}
\subsection{Platforms}

To demonstrate the generalizability of our method, we show our results on three different robot platforms, with our perception running in-the-loop on our primary system and on teleoperated data from the other two. An overview is provided in Table \ref{tab:robot_configs}. For each robot we test a different combination of visual back-end and cost function.

\subsubsection{Full-Scale All-Terrain Vehicle}
We run our autonomy experiments on a Yamaha Viking All-Terrain Vehicle (ATV), first modified by Mai et al. \cite{Mai-2020} and further modified by Sivaprakasam et al. \cite{tartandrive2}. We use the ViT-B version of DINOv2 and use features from a held-out area to compute clusters for the VLAD descriptors.

\begin{table}[b]
    \caption{Robot Autonomy Configurations}
    \centering
    \resizebox{\linewidth}{!}{
    \begin{tabular}{|c||c|c|c|c|c|}
        \hline  %
        \textbf{Robot} & Dynamics & Environment & Depth Sensor & Vis. Backbone & Cost Function\\
        \hline
        ATV & Ackermann & Nature & Lidar & DINOv2 & Roughness\\
        Wheelchair & Skid-Steer & Urban & Stereo Cam. & RADIOv2.5 & Roughness\\ 
        ANYmal & Quadruped & Urb. + Nat. & Lidar & DINOv2 & Velocity Error\\
        \hline
    \end{tabular}
    }
    \label{tab:robot_configs}
\end{table}

\subsubsection{Autonomy-Equipped Wheelchair}
We test our method on data from a wheelchair equipped with a stereo camera driven in an urban environment, and use the ViT-B version of RADIOv2.5 \cite{radio}. We use the same roughness cost function used on the ATV (without the shock-travel sensor information), and one-shot cost augmentation is used to assign cars with high cost.

\subsubsection{ANYmal Quadruped}
We also test our method on an ANYmal quadruped robot developed by ANYbotics \cite{anymal}, using data provided from Wild Visual Navigation (WVN) \cite{mattamala2024wildvisualnavigationfast}. We use the cost function provided in WVN, based on the discrepancy between desired and commanded velocity. One-shot cost augmentation to assigns trees and building walls with high cost. We compute descriptors using the same clusters generated from the off-road site used for the ATV.

\subsection{ATV Autonomous Navigation}

\subsubsection{Baselines}
\begin{figure}
	\centering
 	\includegraphics[width=\linewidth]{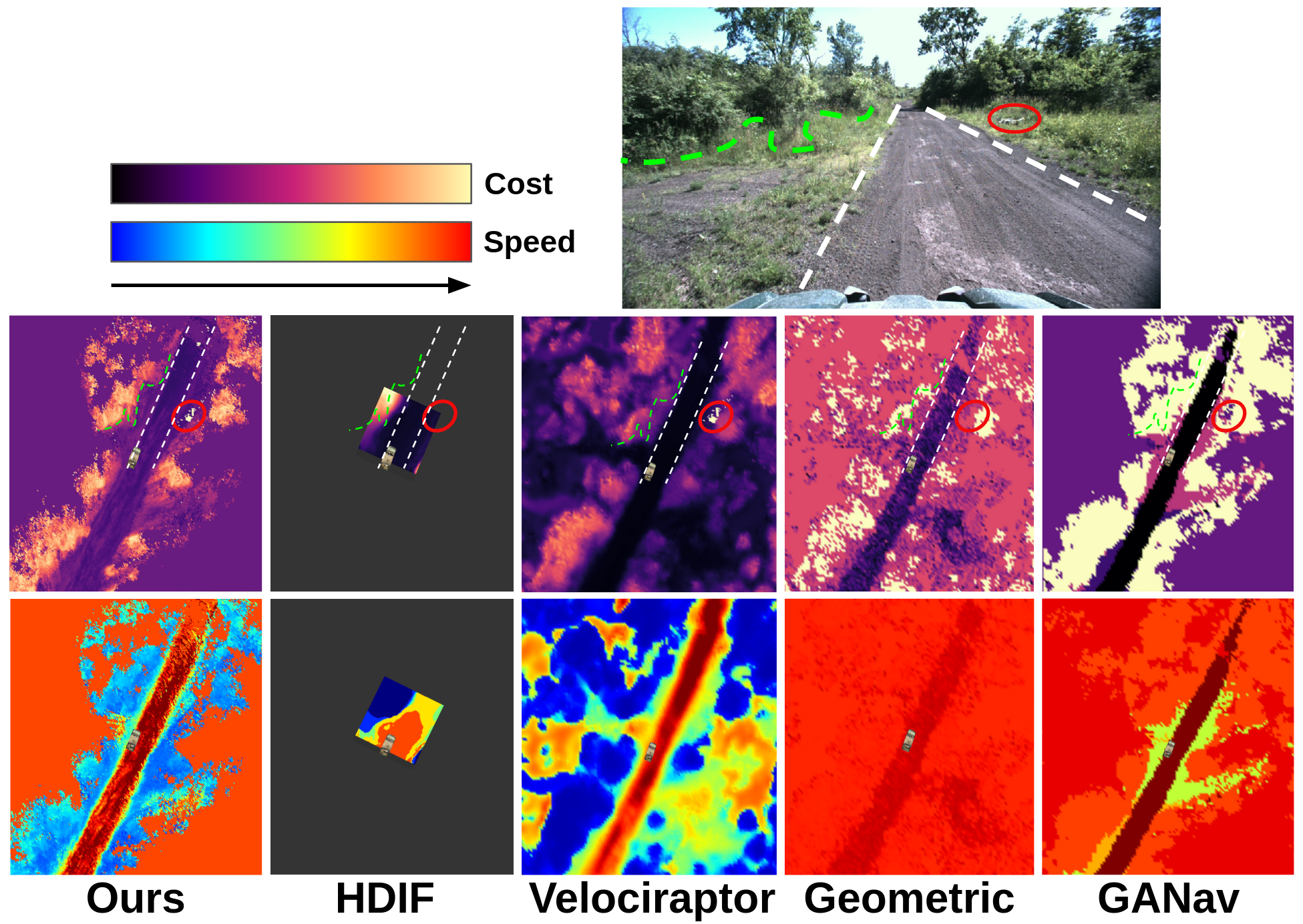}

	\caption{Comparison of our method against baselines. Note our method's ability to distinguish the tree line (green dashed line), trail (white dashed line), and the shattered TV hidden in the bushes (red circle).}
	\label{fig:qualitative_all}
\end{figure}
\begin{figure}[b]
	\centering
	\includegraphics[width=.9\linewidth,height=.45\linewidth]{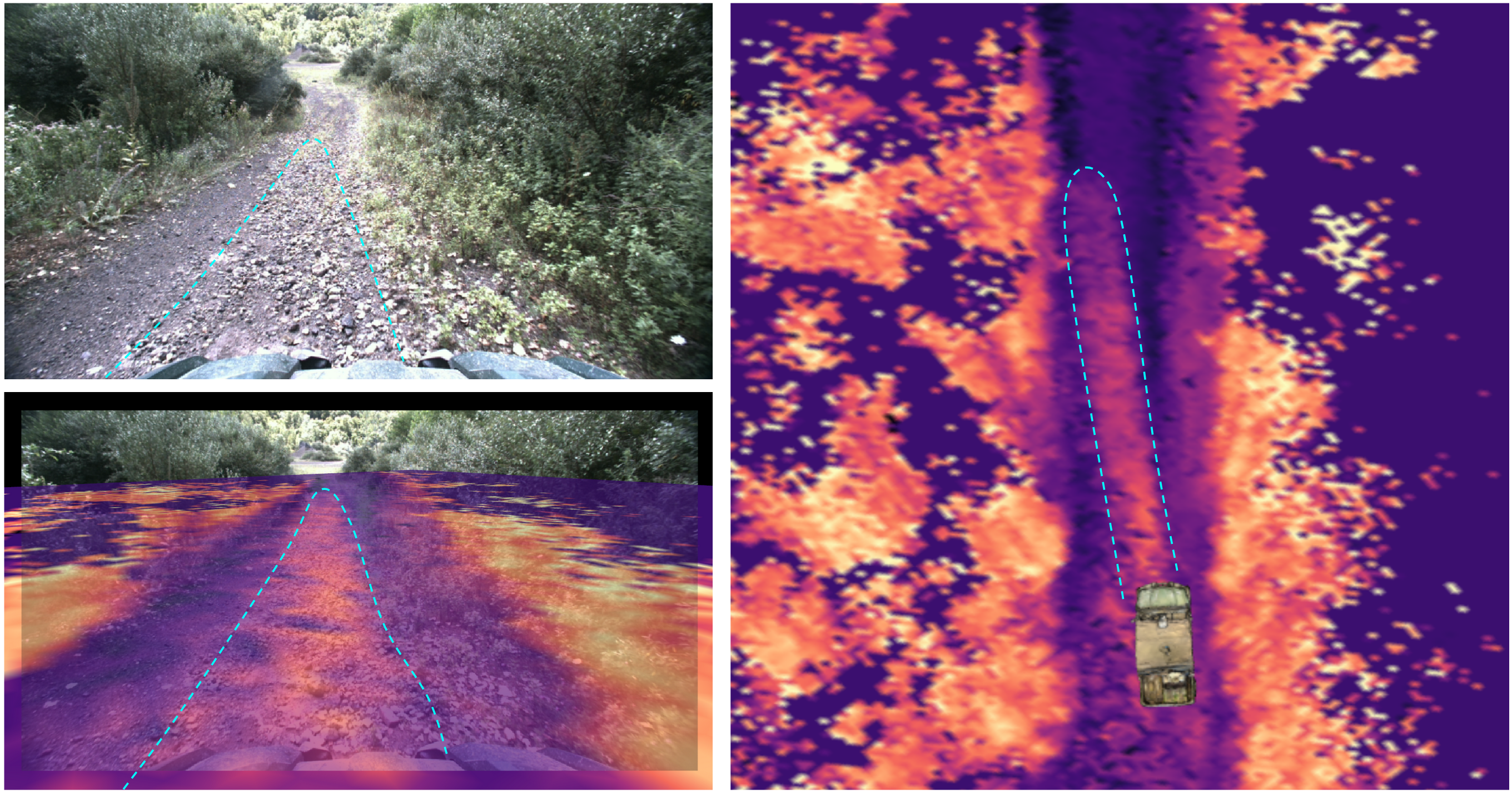}
	\caption{$\SystemName$'s ability to distinguish fine-grained terrain: Rough gravel in the middle of the trail is high cost.}
	\label{fig:turnpike}
\end{figure}

Using an MPPI \cite{mppi} controller to actuate on the costmaps, we compare our method on our primary system against four other baselines:
\begin{itemize}
\item \textbf{Geometric cost function}: we use the cost function used in training ALTER \cite{chen2023learningonthedrive} as a geometry-informed baseline. We compute a speedmap that encourages higher speed in areas with high planarity.
\item \textbf{Semantic segmentation using GANav \cite{ganav}}: we use GANav, mapping the semantic logits using the same pipeline used for our method, and associate the classes to hand-tuned costs and speeds.
\item \textbf{How Does It Feel? (HDIF) \cite{10160856}}: we compare against HDIF as it is a learning-based predecessor to this work, using a constant desired speed and limited range due to runtime constraints.
\item \textbf{Velociraptor \cite{triest2024velociraptor}}: an inverse reinforcement learning approach that leverages geometric and visual cues to predict costmaps and speedmaps.
\end{itemize}

\begin{table}[t]
    \caption{Capability Comparison Against Baselines}
    \centering
    \resizebox{\linewidth}{!}{
    \begin{tabular}{|c||c|c|c|c|c|}
        \hline        
        \textbf{Method} & \thead{Self\\Supervised} & \thead{Online \\ Adaptation} & \thead{Risk\\Aware} & \thead{OOD\\Detection} & \thead{Velocity\\Informed}\\
        \hline
        Geometry & \ding{51} & \ding{55} & \ding{55} & \ding{55} & \ding{55} \\
        GA-Nav & \ding{55} & \ding{55} & \ding{55} & \ding{55} & \ding{55}\\
        HDIF & \ding{51} & \ding{55} & \ding{55}  & \ding{55}& \ding{51}\\
        Velociraptor  & \ding{51} & \ding{55} & \ding{51}  & \ding{51}& \ding{51} \\
        \SystemName  & \ding{51} & \ding{51} & \ding{51}  & \ding{51}& \ding{51} \\
        \hline
    \end{tabular}
    }
    \label{tab:baseline_compare}
\end{table}

\begin{figure}[b]
	\centering
	\includegraphics[width=.99\linewidth]{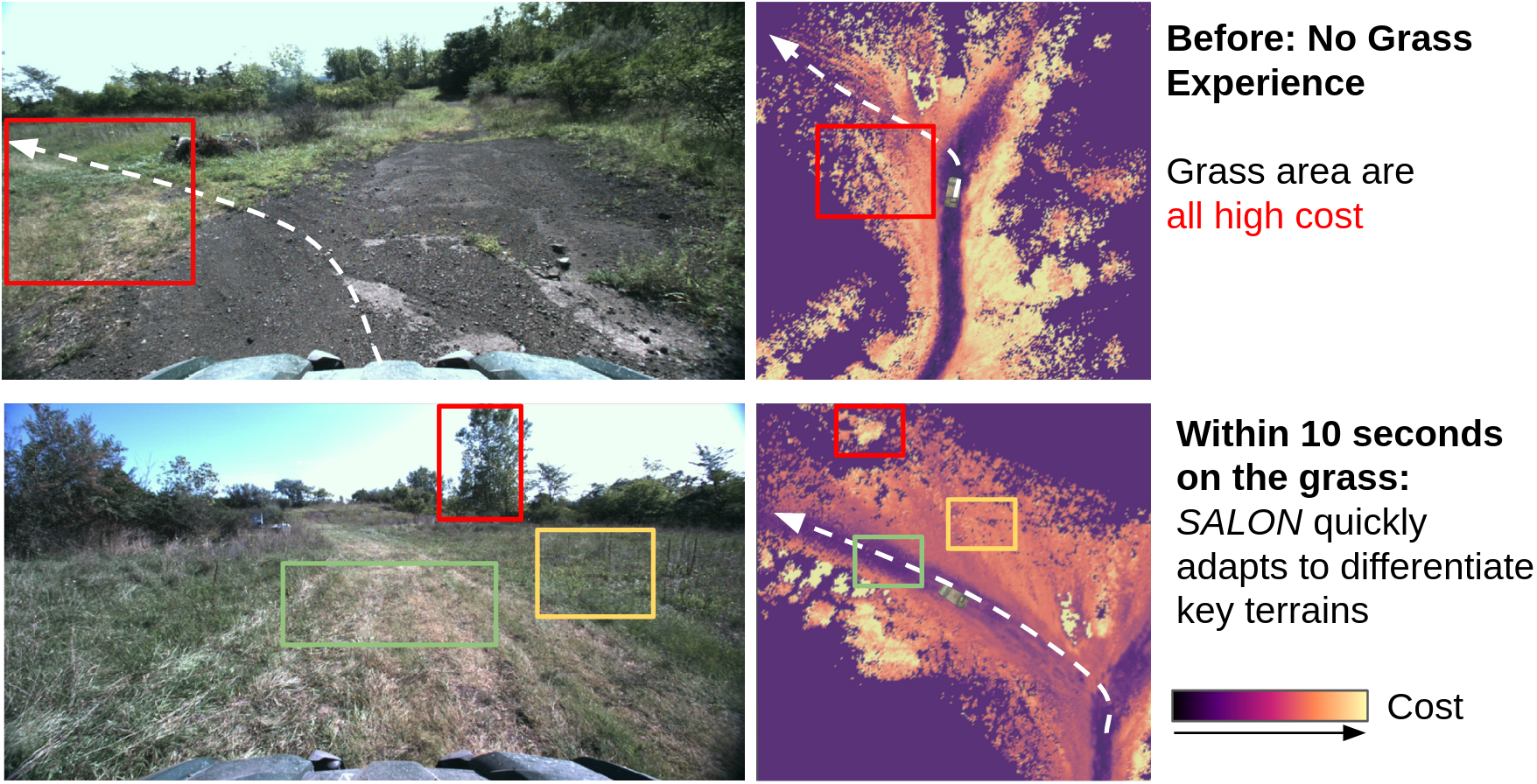}
     \caption{Example of $\SystemName$'s fast adaptation: Within 10 seconds of experiencing grass, $\SystemName$ is able to quickly differentiate key terrains, such as, ideal short grass, riskier vegetation and lethal trees.}
	\label{fig:adaptation_fig8}
\end{figure}

\subsubsection{Qualitative Results}
Within less than 60 seconds of experience, our method appears to outperform the simpler baselines (GANav and geometric cost function) and predict maps at a level of detail more similar to Velociraptor (Fig. \ref{fig:qualitative_all}). Note that Velociraptor also uses geometric features from lidar as an input, enabling more range and a wider field-of-view.

\definecolor{Gray}{gray}{0.9}
\begin{table*}[h]
\centering
\caption{Navigation Performance Metrics}
\label{tab:nav-metrics}
\begin{tabular}{|c|ccccc|ccccc|ccccc|ccccc|}
\hline
\textbf{Method} & \multicolumn{5}{c|}{\textbf{\# Interactions}} & \multicolumn{5}{c|}{\textbf{\# Undesirable Behavior}} & \multicolumn{5}{c|}{\textbf{Avg. Speed (m/s)}} & \multicolumn{5}{c|}{\textbf{Avg. Roughness}} \\ \hline
\rowcolor{Gray} Lap Number                & 1       & 2       & 3       & 4      & 5      & 1         & 2        & 3        & 4        & 5        & 1       & 2       & 3       & 4       & 5      & 1       & 2       & 3      & 4      & 5      \\ \hline
Geometry        & 3       & 6       & 5       & 5      & 4      & 1         & 1        & 2        & 1        & 2        & \textbf{5.2}       & 4.5       & 4.5       & 4.6       & 4.6      & .30       & .26       & .27      & .27      & .27      \\ \hline
GANav           & 4       & 6       & 4       & 4      & 4      & 1         & 2        & 2        & 2        & 2        & 3.7       & 3.5       & 3.5       & 3.7       & 3.6      & .18       & .19       & .18      & \textbf{.19}      & .19      \\ \hline
HDIF            & 3       & 6       & 3       & 3      & 4      & \textbf{0}         & 1        & 2        & 1        & \textbf{0}        & 4.2       & 3.6       & 4.0       & 4.2       & 3.8      & .23       & .21       & .24      & .23      & .23      \\ \hline
Velociraptor          & 1       & \textbf{0}       & 1       & 1      & 1      & 1         & 2        & 2        & 1        & 2  & 5.1       & \textbf{4.7}       & 4.7       & \textbf{4.7}       & \textbf{4.9}      & .26       & .26       & .28      & .27      & .28      \\ \hline
\SystemName, $R_{max}=.2$   & 2       & 1       & \textbf{0}       & \textbf{0}      &\textbf{ 0}      & \textbf{0}         & 1        & \textbf{0}        & 1        & \textbf{0}  & 3.4       & 3.6       & 3.4       & 3.7       & 3.9      & \textbf{.16}       & \textbf{.17}       & \textbf{.17}      & \textbf{.19}      & \textbf{.18}      \\ \hline
\SystemName, $R_{max}=.4$   & 2       & \textbf{0}       & 1       & \textbf{0}      & 1      & \textbf{0}         & \textbf{0}        & 1        & 1        & \textbf{0}  & 3.8       & 4.5       & \textbf{4.8}       & 4.6       & 4.5      & .19       & .25       & .25      & .24      & .24      \\ \hline
\end{tabular}
\end{table*}

We find that after adaptation, our predicted maps reflect expected behavior conditioned on various velocities and max roughness values (Fig. \ref{fig:adaptation_title_horizontal}). We also present additional qualitative examples that highlight our method's ability to not only distinguish high-level terrain types --- such as trees, grass, and trail --- but also pick up on visual cues such as grass density and gravel in order to make expressive predictions (Fig. \ref{fig:turnpike}).

To better observe the adaptive behavior of our method, we show predictions less than 10 seconds apart in the same lap (Fig. \ref{fig:adaptation_fig8}). Without any experience on grass, our method can pick out trees and trail, but believes that the grass is a higher cost than it should be.  After driving on it for just a few seconds it has a finer understanding of the terrain, able to distinguish between smooth and rough vegetation.

\subsubsection{Quantitative Results}
We run all baselines as well as our own method for five laps in a "figure 8" style course with 50m waypoint spacing, evaluated on four metrics (Table \ref{tab:nav-metrics}). We count the number of times the safety driver intervened in order to prevent damage, as well as the number of "undesirable behaviors" where the driver didn't need to intervene but the system could have taken a better route (for example driving through a rough patch of grass when there is a trail right next to it). 

Our method outperforms the baselines in all metrics apart from average speed. While Velociraptor had a higher average speed it also had more interventions, some of which were a result of driving too fast. Due to the fragility of our vehicle we prioritize number of interventions over speed, but we recognize that there is an inherent trade-off between the two and preferences may vary based on robot and operator.

\subsection{Qualitative Comparison to State-of-the-Art Adaptation Method}
\begin{figure}[]
	\centering
	\includegraphics[width=.99\linewidth]{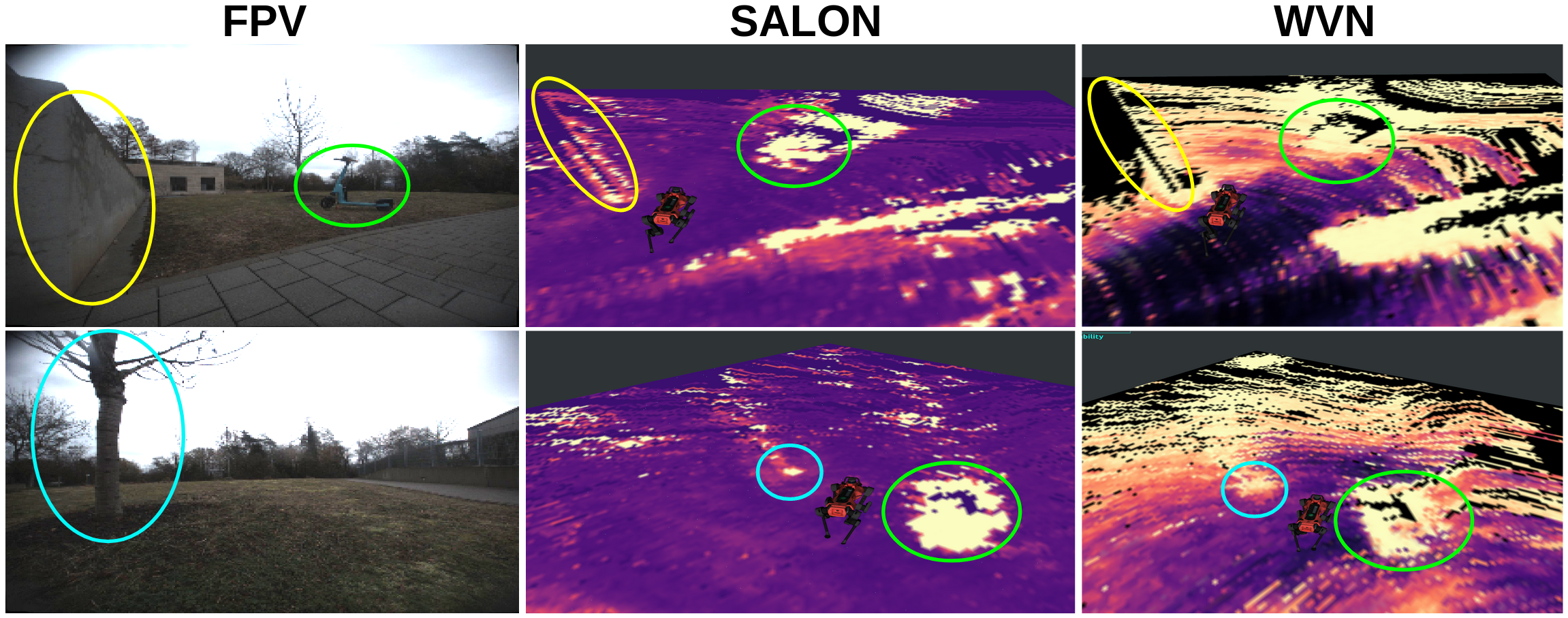}
    \caption{With the same amount of data as WVN, our method is able to correctly cost lethal objects like trees and walls without incorrectly costing short grass.}
	\label{fig:wvn_compare}
\end{figure}

We test our method on the quadruped used in WVN. To compare the two methods qualitatively, we project the FPV cost predictions from WVN into a costmap using the same mapping pipeline in our approach. Our method appears to be more consistent as shown in Fig. \ref{fig:wvn_compare}, likely due to the fact that our method makes predictions in the map space, in contrast to than mapping predictions made in the FPV space. Additionally, while WVN appears to cost grass more heavily than pavement compared to ours we find that this is not representative of their cost function, which considers the two types of terrain to be similar.

\subsection{Qualitative Evaluation in Urban Environment}
\begin{figure}[]
	\centering
	\includegraphics[width=.85\linewidth]{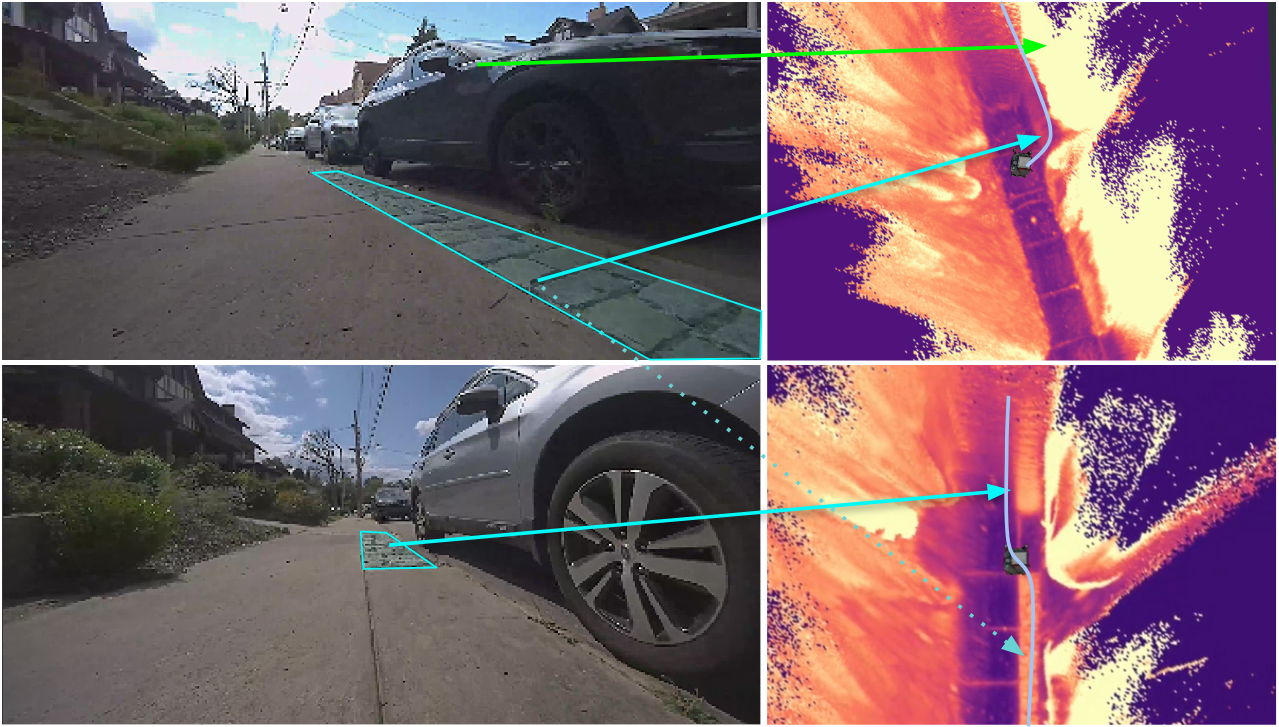}
	\caption{Evaluation on urban wheelchair: After driving over rough cobblestone, the system quickly recognizes within 5 seconds that it is much rougher than the smooth sidewalk.}
	\label{fig:wheelchair}
\end{figure}

Using the same cost function used on the ATV but a different visual backend (RADIOv2.5), we run our method on a wheelchair on sidewalks in an urban environment. In Fig. \ref{fig:wheelchair}, we highlight a scenario where initially, the robot incorrectly equates rough cobblestone with smooth pavement. After driving over the rough section for a few seconds, it learns to distinguish between the two in the environment ahead of it. We also find that it is able to pick up on fine details such as cracks in the sidewalk.

\section{Conclusions and Future Work}
In this work we present $\SystemName$, a framework for predicting costmaps and speedmaps that allow a system to adapt in real time to novel experiences by associating generalizable features from visual foundation models with proprioceptive signals. This results in a system that can make more nuanced predictions about its environment than the prior state-of-the-art that in turn allow for improved navigation behaviors. Further, we demonstrate promising results on multiple robots to highlight the generalizability of our method.

While we improve on many of the issues presented in our baselines, there still remains a number of future directions. For example, the exploration of the trade-offs between different distribution assumptions would be highly beneficial. Additionally, our strategy for speedmap prediction assumes that roughness increases relatively monotonoically with speed. While we observe this to be mostly the case in our system, there could exist other systems built in a way such that certain terrain is actually less rough at high speeds.

There is also a need for unified benchmarking of off-road autonomy as a whole. While signals such as interventions and speed are certainly strong identifiers of success, they often have an inverse relationship with the weight of each being ambiguous due to vehicle and user constraints.

\section*{Acknowledgement}
We thank Humeyra Kacar and Anton Yanovich for their assistance in field testing. We'd also like to thank Nathan Litzinger, Bangjie Xue, Victor Zayakov, and Jiahe Xu for their work on the setting up the wheelchair testing platform for data collection.

{
\bibliographystyle{IEEEtran}
\bibliography{refs}
}

\end{document}